\documentclass[11pt,a4paper]{article}
\usepackage[hyperref]{acl2021}
\usepackage{times}
\usepackage{latexsym}
\usepackage{adjustbox}
\usepackage{graphicx}
\usepackage{multirow}

\usepackage{microtype}

\aclfinalcopy 

\title{Human Evaluation of Creative NLG Systems: An Interdisciplinary Survey on Recent Papers}

\author{Mika Hämäläinen \\
  Faculty of Arts \\
  University of Helsinki \\
  \texttt{mika.hamalainen@helsinki.fi} \\\And
  Khalid Alnajjar\\
  Faculty of Arts \\
  University of Helsinki \\
  \texttt{khalid.alnajjar@helsinki.fi} \\}

\date{}

\begin{document}
\maketitle
\begin{abstract}
We survey human evaluation in papers presenting work on creative natural language generation that have been published in INLG 2020 and ICCC 2020. The most typical human evaluation method is a scaled survey, typically on a 5 point scale, while many other less common methods exist. The most commonly evaluated parameters are meaning, syntactic correctness, novelty, relevance and emotional value, among many others. Our guidelines for future evaluation include clearly defining the goal of the generative system, asking questions as concrete as possible, testing the evaluation setup, using multiple different evaluation setups, reporting the entire evaluation process and potential biases clearly, and finally analyzing the evaluation results in a more profound way than merely reporting the most typical statistics.
\end{abstract}

\section{Introduction}

Human evaluation in natural language generation (NLG) has become a hot topic lately, with the emergence of several survey papers on the topic that study how human evaluation has been conducted in the past in the field of NLG in general \cite{howcroft-etal-2020-twenty,belz-etal-2020-disentangling}. This has led to several recent evaluation frameworks for evaluating the output of NLG systems \cite{liu2020glge,gehrmann2021gem}.

However, not all natural language generation tasks are of the nature that they are designed to convey factual information. Some of the NLG tasks deal with producing text of aesthetic nature such as poetry, stories, humor and so on. We call these creative NLG tasks. These types of tasks are simultaneously researched in two distinct fields of science: natural language processing (NLP) and computational creativity (CC). Existing survey papers have only focused on NLP research and they have not made a distinction between creative and non-creative NLG.

NLP and CC fields conduct work from very different starting points \cite{ws-2016-inlg}. NLP is often state-of-the-art driven whereas CC presents more of exploratory research without pursuing scores that outperform a baseline. In this paper, we want to study how human evaluation of creative NLG systems is conducted in the world of NLP and in the world of CC, what similarities there are and whether the two fields can learn something from each other.

We base our research on a literature review on the papers dealing with human evaluated creative NLG published in the 2020 editions of the International Conference on Computational Creativity (ICCC) and of the International Conference on Natural Language Generation (INLG). We picked these conferences as ICCC is the most important venue for CC research, and INLG the most important NLP focused venue for NLG research.

Our results show that there is no consensus at the moment on how evaluation should be conducted despite the many different efforts of establishing guidelines for evaluating computationally creative output \cite{pease2011impact,jordanous2012standardised,lamb2018evaluating,hamalainen2020generating}. We reflect on the results of our survey and propose a road-map for more sound future evaluation practices.

\begin{table*}[ht]
\centering
\renewcommand{\tabcolsep}{2pt}
\begin{adjustbox}{max width=\textwidth}
\begin{tabular}{|l|l|l|l|l|}
\hline
Paper                                                               & NLG task                                                                                               & Evaluated parameters                                                                                                                                                                                              & Questions motivated                                                                                     & Evaluation type                                                                                                                \\ \hline
1. \citealt{mathewson2020shaping}                     & Collaborative dialogue                                                                                 & engagement                                                                                                                                                                                                        & \begin{tabular}[c]{@{}l@{}}Engagement measured the \\ notions of revealing and concealing.\end{tabular} & Ranking models                                                                                                                 \\ \hline
2. \citealt{cheatley2020co}                           & Song writing tool                                                                                      & \begin{tabular}[c]{@{}l@{}}Support of self-expression, therapeutic value\\ and receptiveness to the tool and songs created\end{tabular}                                                                           & Not discussed                                                                                           & User study (qualitative)                                                                                                       \\ \hline
3. \citealt{mirowski2020rosetta}                      & \begin{tabular}[c]{@{}l@{}}Auxiliary tool for \\ improv theater\end{tabular}                           & Based on critics' previews and reviews                                                                                                                                                                            & No questions                                                                                            & public performance                                                                                                             \\ \hline
4. \citealt{spendlove2020creating}                    & \begin{tabular}[c]{@{}l@{}}Generating six \\ word stories\end{tabular}                                 & coherence, impactfulness                                                                                                                                                                                          & Not discussed                                                                                           & 5 point scale                                                                                                                  \\ \hline
5. \citealt{ammanabrolu2019toward}                    & \begin{tabular}[c]{@{}l@{}}Quest generation in \\ text adventure games\end{tabular}                    & \begin{tabular}[c]{@{}l@{}}coherence, originality (novelty), sense of \\ acomplishment (value), unpredictability (surprise)\end{tabular}                                                                          & By Boden's theory on creativity                                                                         & 7 point scale                                                                                                                  \\ \hline
6. \citealt{mendes2020comparing}                      & \begin{tabular}[c]{@{}l@{}}Headline-proverb \\ pair generation\end{tabular}                            & relatedness, funniness                                                                                                                                                                                             & Not discussed                                                                                           & 4 point scale                                                                                                                  \\ \hline
7. \citealt{tylercomputational}                       & Pun generation                                                                                         & \begin{tabular}[c]{@{}l@{}}funniness, surprise, cleverness, did the user laugh,\\ wit, ingenuity, timelessness, and accessibility\end{tabular}                                                                    & Not discussed                                                                                           & 5 point scale                                                                                                                  \\ \hline
8. \citealt{mendes2020teco}                           & \begin{tabular}[c]{@{}l@{}}Contextual headline \\ adaptation\end{tabular}                              & syntax, relatedness, funniness                                                                                                                                                                                    & Not discussed                                                                                           & 3 point scale                                                                                                                  \\ \hline
9. \citealt{hamalainen2020automatic} evaluation 1      & \multirow{2}{*}{\begin{tabular}[c]{@{}l@{}}Dialectal adaptation\\ of generated poetry\end{tabular}}    & \begin{tabular}[c]{@{}l@{}}poem (yes/no), typicality, understandability, \\ quality of language, evoked imagery, \\ evoked emotions, annotator's liking\end{tabular}                                              & Previous research                                                                                       & 5 point scale                                                                                                                  \\ \cline{1-1} \cline{3-5} 
 \citealt{hamalainen2020automatic} evaluation 2      &                                                                                                        & \begin{tabular}[c]{@{}l@{}}emotivity, originality, creativity, \\ poem-likeness, artificiality, fluency\end{tabular}                                                                                              & Not discussed                                                                                           & Association                                                                                                                    \\ \hline
10. \citealt{saveryshimon}                             & \begin{tabular}[c]{@{}l@{}}Real time human-\\ machine rap battles\end{tabular}                         & \begin{tabular}[c]{@{}l@{}}annotator's perception,  coherence, rhythm, \\ rhyme, quality, enjoyment, relation between \\ the hip hop and metal dataset, \\ and relationship between input and output\end{tabular} & By research questions                                                                                   & \begin{tabular}[c]{@{}l@{}}open ended questions \\ + automatic analysis, \\ preference\end{tabular}                            \\ \hline
11. \citealt{hugoweird}                                & \begin{tabular}[c]{@{}l@{}}Song lyric \\ transformation\end{tabular}                                   & \begin{tabular}[c]{@{}l@{}}familiarity, novelty, grammaticality, \\ semantics, singability, overall appreciation \\ and topicality\end{tabular}                                                                   & Not discussed                                                                                           & \begin{tabular}[c]{@{}l@{}}5 point scale and\\ picking the most suitable topic\end{tabular}                                    \\ \hline
12. \citealt{shihadehemily}                            & \begin{tabular}[c]{@{}l@{}}Emily Dickinson style\\ poem generation\end{tabular}                        & \begin{tabular}[c]{@{}l@{}}typicality, understandability, \\ quality of language, evoked imagery, \\ evoked emotions, annotator's liking\end{tabular}                                                             & Previous research                                                                                       & 5 point scale                                                                                                                  \\ \hline
13. \citealt{gong-etal-2020-rich}                      & Text style transfer                                                                                    & \begin{tabular}[c]{@{}l@{}}content preservation, transfer strength\\ and fluency\end{tabular}                                                                                                                     & Automated evaluation                                                                                    & picking the best                                                                                                               \\ \hline
14. \citealt{obeid-hoque-2020-chart}                   & \begin{tabular}[c]{@{}l@{}}Text generation \\ from charts\end{tabular}                                 & \begin{tabular}[c]{@{}l@{}}informativeness, conciseness, \\ coherence, fluency, factuality\end{tabular}                                                                                                           & Not discussed                                                                                           & \begin{tabular}[c]{@{}l@{}}5 point scale and \\ yes/no/partially/can't decide \\ for factuality\end{tabular}                   \\ \hline
15. \citealt{lee-2020-stable}                          & Style transform                                                                                        & content, fluency, and style                                                                                                                                                                                        & Not discussed                                                                                           & 5 point scale                                                                                                                  \\ \hline
16. \citealt{mendes-goncalo-oliveira-2020-amplifying}  & \begin{tabular}[c]{@{}l@{}}Enhancing headlines \\ with creative expressions\end{tabular}               & relatedness, funniness                                                                                                                                                                                             & Not discussed                                                                                           & 4 point scale                                                                                                                  \\ \hline
17. \citealt{langner-2020-omega}                       & \begin{tabular}[c]{@{}l@{}}Referring expression \\ generation in a \\ virtual environment\end{tabular} & \begin{tabular}[c]{@{}l@{}}comprehension based on \\ identification time, error rate \\ and repetition counts\end{tabular}                                                                                        & Not discussed                                                                                           & \begin{tabular}[c]{@{}l@{}}user study based on \\ quantitative values\end{tabular}                                             \\ \hline
18. \citealt{scialom-etal-2020-bert}                   & \begin{tabular}[c]{@{}l@{}}Question generation \\ from images\end{tabular}                             & \begin{tabular}[c]{@{}l@{}}readability, caption relevance \\ and image relevance\end{tabular}                                                                                                                     & Not discussed                                                                                           & 5 point scale                                                                                                                  \\ \hline
19. \citealt{ilinykh-dobnik-2020-image}                & \begin{tabular}[c]{@{}l@{}}Multi-sentence image\\ description generation\end{tabular}                  & \begin{tabular}[c]{@{}l@{}}word choice, object salience, \\ sentence structure and paragraph coherence\end{tabular}                                                                                               & Not discussed                                                                                           & slider                                                                                                                         \\ \hline
20. \citealt{akermi-etal-2020-tansformer}              & Question answering                                                                                     & relevance, errors                                                                                                                                                                                                 & Not discussed                                                                                           & \begin{tabular}[c]{@{}l@{}}relevance (correct/not correct), \\ error type checkboxes, \\ open ended comment field\end{tabular} \\ \hline
21. \citealt{nikolov-etal-2020-rapformer} evaluation 1 & \multirow{3}{*}{Rap lyric generation}                                                                  & style, meaning, familiarity                                                                                                                                                                                       & Not discussed                                                                                           & 5 point scale                                                                                                                  \\ \cline{1-1} \cline{3-5} 
 \citealt{nikolov-etal-2020-rapformer} evaluation 2 &                                                                                                        & Turing test                                                                                                                                                                                                       & Not discussed                                                                                           & \begin{tabular}[c]{@{}l@{}}picking which out of 2 \\ is written by a human\end{tabular}                                        \\ \cline{1-1} \cline{3-5} 
 \citealt{nikolov-etal-2020-rapformer} evaluation 3 &                                                                                                        & Turing test                                                                                                                                                                                                       & Not discussed                                                                                           & human written (yes/no)                                                                                                         \\ \hline
22. \citealt{wang-etal-2020-reviewrobot}               & Paper review generation                                                                                & constructivenness and validity                                                                                                                                                                                    & Not discussed                                                                                           & not stated                                                                                                                     \\ \hline
23. \citealt{hedayatnia-etal-2020-policy}              & \begin{tabular}[c]{@{}l@{}}Response generation\\ in a dialog system\end{tabular}                       & appropriateness                                                                                                                                                                                                   & Previous research                                                                                       & picking the best                                                                                                               \\ \hline
\end{tabular}
\end{adjustbox}
\caption{Evaluated parameters, their motivation and evaluation type in the surveyed papers}
\label{tab:evaluation-info}
\end{table*}

\section{Surveying human evaluation methods}

In this section, we go trough how human evaluation was conducted in the papers we selected for the survey. From the ICCC proceedings, we included all papers that dealt with NLG and had a human evaluation. We did not survey papers that presented work on generating something else than language such as music. From the INLG proceedings, we picked all papers that presented work on an open-ended NLG problem the output of which could exhibit some creativity ruling out papers that dealt with purely factual data-to-text generation tasks.

In the ICCC 2020, there were 12 papers that presented human evaluated work on creative NLG, and in the INLG 2020, there were 11 such papers. We selected these papers for our survey. Fortunately, both of the venues had relatively the same amount of papers.

When surveying the papers, we only focused on human evaluation and we wanted to know what the NLG task was, what parameters were being evaluated (usually reflected by the evaluation questions), how these parameters (questions) were motivated and how the actual evaluation was conducted methodologically. We also paid attention to the evaluation setup: the number of evaluators and samples used and whether the evaluators were experts or laymen. Finally, we looked into the discussions and conclusions presented in the papers to see what role the human evaluation had there, especially in relation to concrete future directions in improving the system based on the evaluation results.

\subsection{What is evaluated?}

Table \ref{tab:evaluation-info} shows the results of our survey in terms of what parameters were evaluated and how the evaluation was conducted. Papers 1-12 were published in ICCC and represent the CC field, whereas papers 13-23 were published in INLG representing the NLP side of the same coin.


When looking at the results, we can immediately see that there is quite a range of different NLG tasks. Even for papers that deal with very similar tasks such as papers 2, 10, 11 and 21, the framing of the problem is very different ranging from lyric transformation to full-blown human versus computer rap battles. The evaluated parameters were also very different.

Despite the parameters being very different from each other, several papers evaluated \textbf{meaning} in one way or another, for example, papers 4, 5, 10, 14 and 19 evaluated coherence, paper 11 semantics and paper 21 meaning.  Papers 9 and 12 evaluated understandability, which is not directly the same as meaning.

\textbf{Syntactic correctness} of the language was also one of the commonly evaluated features. Papers 9, 13 and 14 measure fluency, paper 11 grammaticality, papers 9 and 12 quality of language and paper 8 syntax. In addition paper 18 evaluated readability, which is partially related to correctness and partially to meaning.

One of the parameters that was evaluated through multiple synonyms and even antonyms was \textbf{novelty}. Papers 5 and 9 evaluated originality, paper 11 novelty, paper 7 surprise and paper 5 unpredictability. Papers 9 and 12 evaluated the opposite of novelty, which is typicality.

\textbf{Relevance} was also commonly evaluated in papers 18 and 20. The parameter was evaluated as relatedness in papers 6, 8 an 16, although all of them are by the same authors.

Many papers also evaluated \textbf{emotional value}. Such as paper 9 through emotivity, paper 10 through enjoyment, paper 11 through engagement, papers 9 and 12 through evoked emotions and papers 7, 6, 8 and 16 through funniness, although three of these papers were by the same authors.

\subsection{Why are the evaluation parameters chosen?}

The aforementioned parameters do not cover all the parameters that were used in evaluation, however, they were the most typical ones. When we look into how the evaluation parameters were selected, we can notice that most of the papers do not present any reasoning as to why these are the relevant attributes to look at. 

The few papers that did present a reasoning, had many different reasons for the evaluated parameters. Paper 1 motivates the evaluated parameter by stating that it evaluates revealing and concealing parameters that were defined important for the task. Paper 3 did not have any parameters at all for evaluation. Paper 5 motivated the evaluated parameters through an existing theory on computational creativity \cite{boden2007creativity}. Paper 10 had formulated the evaluated parameters based on the research questions established in the paper. Paper 13 formulated the evaluated parameters so that they would measure the same things as their automated evaluation.

Paper 9 and 12 used evaluation questions originally established by \citet{toivanen2012corpus}. While basing evaluation on existing research makes the evaluation questions sound more well motivated, the original paper where these evaluation questions were first established did not present any reasoning as to why these should be the evaluation questions to be used with generated poetry. Also paper 23 stated they used "a similar setup" as proposed by \citet{li2016deep}. In practice this meant that whereas the original paper proposed 3 different evaluation setups, paper 23 only presented one of them. The reasoning for this evaluation was not discussed in the original paper.

\subsection{How is the evaluation conducted?}

Most of the papers present only one human evaluation method. The exceptions are paper 9 that presents two distinct evaluation setups and paper 21 that presents 3 distinct evaluation setups.

The most common way of conducting a human evaluation is to use a questionnaire that is rated on a \textbf{numerical scale}. Papers 4, 7, 9 (evaluation 1) 11, 12, 14, 15, 18 and 21 (evaluation 1) used a 5 point scale. Papers 6 and 16, written by the same authors, use a 4 point scale, and paper  8, also by the same authors, uses a 3 point scale. Paper 7 uses as big as a 7 point scale. The most deviant one of the papers using a numeric scale is paper 19. This paper presents a continuous slider the annotators can move freely. Some of the papers use a different scale for one of the questions.

The second most typical evaluation method is based on \textbf{preference}. Here the outputs are preferred or ranked in relation to each other. Paper 1 presents a ranking method where different models are ranked based on which one is the best. Paper 9 (evaluation 2) presents two poems side by side and asks annotators to associate the presented parameters with either one of them. Paper 10 uses preference of output as one of the evaluation criteria. Papers 13 and 23 ask the annotators to pick the best output candidate. Paper 21 (evaluation 2) asks the annotators to guess which output is human written and which AI written. Paper 11 asks the annotators to rank the most suitable topic. This is slightly different as here the annotators are not asked to rank the output per se. As we can see, there are a great number of different variations in how this type of an evaluation is conducted. As opposed to the most popular evaluation method, these methods only give relative results. This means that even if all of the output was bad, one of them is still picked as the best.

Two papers, 2 and 17, present a user-study. Paper 2 conducts this in a qualitative way with open ended questions where the discussion is directed towards the parameters that the authors wanted to measure. The discussions with the participants are not fully reported in the paper, instead the authors present some quotes relating to the parameters in study in a non-rigorous fashion. Paper 17 presents a quantitative user-study where the results are analyzed based on different values such as execution time that were gathered during the user-study.

Paper 3 presents something completely unique in terms of evaluation. The authors organize live improv theater sessions with the system and base the results on the reviews and previews by critics. However, these were not discussed in the paper in detail, but rather some cherry picked quotations were reported.

Paper 10 was another paper to conduct a qualitative evaluation. The annotators were asked to answer to open-ended questions. The input from the annotators was then automatically processed to reach to conclusions. An open-ended comments field was also provided in paper 20, however, the paper focused on discussing the results of the two other questions in the questionnaire. The annotators were asked to give a binary rating on whether the output was relevant or not, similarly, paper 9 (evaluation 1) presented one binary question about poeticity and Paper 21 (evaluation 3) presented a binary question whether the output was human authored. In addition, paper 20 asked the annotators to indicate which types of errors the output had by providing a set of check-boxes with predefined error types.

Unlike the rest of the papers, paper 22 did not explain how the evaluation was conducted in any detail. The results were percentages, which indicates that the evaluation might have been based on binary questions.

\begin{table*}[ht]
\centering
\renewcommand{\tabcolsep}{2pt}
\begin{adjustbox}{max width=\textwidth}
\begin{tabular}{|l|l|l|l|}
\hline
Paper                                                               & Experts                                                                 & Number of annotators                                      & Number of samples                                                                                                                        \\ \hline
1. \citealt{mathewson2020shaping}                     & yes                                                                     & 4                                                         & \begin{tabular}[c]{@{}l@{}}3 conversations \\ (5 utterance-response\\ pairs in each)\end{tabular}                                        \\ \hline
2. \citealt{cheatley2020co}                           & no                                                                      & 3                                                         & \begin{tabular}[c]{@{}l@{}}Free engagement \\ with the system\end{tabular}                                                               \\ \hline
3. \citealt{mirowski2020rosetta}                      & \begin{tabular}[c]{@{}l@{}}yes (reviews), \\ no (audience)\end{tabular} & multiple                                                  & Performance                                                                                                                              \\ \hline
4. \citealt{spendlove2020creating}                    & no                                                                      & 14 per story                                              & 15 stories                                                                                                                               \\ \hline
5. \citealt{ammanabrolu2019toward}                    & no                                                                      & 15 for each game                                          & 2 room layouts                                                                                                                           \\ \hline
6. \citealt{mendes2020comparing}                      & no                                                                      & 4 per headline                                            & 60 headlines                                                                                                                             \\ \hline
7. \citealt{tylercomputational}                       & no                                                                      & 10 in total                                               & \begin{tabular}[c]{@{}l@{}}10 best manually \\ selected puns\end{tabular}                                                                \\ \hline
8. \citealt{mendes2020teco}                           & no                                                                      & 2 in total                                                & 30 headlines                                                                                                                             \\ \hline
9. \citealt{hamalainen2020automatic} evaluation 1      & no                                                                      & 5 per poem variant                                        & 10 poems                                                                                                                                 \\ \hline
\citealt{hamalainen2020automatic} evaluation 2      & no                                                                      & 5 per dialectal-standard Finnish poem pair                & 10 parallel poems                                                                                                                        \\ \hline
10. \citealt{saveryshimon}                             & no                                                                      & 33                                                        & \begin{tabular}[c]{@{}l@{}}1 video clip, \\ hand picked best output, \\ 10 additional video clips \\ and 10 generated tasks\end{tabular} \\ \hline
11. \citealt{hugoweird}                                & no                                                                      & 3 per lyric                                               & 120 lyics                                                                                                                                \\ \hline
12. \citealt{shihadehemily}                            & no                                                                      & 17 in total                                               & \begin{tabular}[c]{@{}l@{}}10 generated +\\  2 Emily Dickinson's poems\end{tabular}                                                      \\ \hline
13. \citealt{gong-etal-2020-rich}                      & no                                                                      & 2 in total                                                & outputs for 100 inputs                                                                                                                   \\ \hline
14. \citealt{obeid-hoque-2020-chart}                   & no                                                                      & 3 per statistic                                           & output for 40 charts                                                                                                                     \\ \hline
15. \citealt{lee-2020-stable}                          & no                                                                      & 6 people per sample                                       & 250 samples                                                                                                                              \\ \hline
16. \citealt{mendes-goncalo-oliveira-2020-amplifying}  & no                                                                      & 4 per headline                                            & 60 headlines                                                                                                                             \\ \hline
17. \citealt{langner-2020-omega}                       & no                                                                      & 34 participants                                           & 10 fixed sessions                                                                                                                        \\ \hline
18. \citealt{scialom-etal-2020-bert}                   & no                                                                      & 3 in total                                                & 50 images                                                                                                                                \\ \hline
19. \citealt{ilinykh-dobnik-2020-image}                & no                                                                      & 154 in total (a participant could rate at most 30 images) & 250 images                                                                                                                               \\ \hline
20. \citealt{akermi-etal-2020-tansformer}              & no                                                                      & 20 in total                                               & 150 questions                                                                                                                            \\ \hline
21. \citealt{nikolov-etal-2020-rapformer} evaluation 1 & yes                                                                     & 3 in total                                                & 100 verses                                                                                                                               \\ \hline
\citealt{nikolov-etal-2020-rapformer} evaluation 2 & yes                                                                     & 3 in total                                                & 100 verses                                                                                                                               \\ \hline
\citealt{nikolov-etal-2020-rapformer} evaluation 3 & yes                                                                     & 3 in total                                                & 100 verses                                                                                                                               \\ \hline
22. \citealt{wang-etal-2020-reviewrobot}               & yes                                                                     & 2 in total                                                & 50 papers                                                                                                                                \\ \hline
23. \citealt{hedayatnia-etal-2020-policy}              & no                                                                      & 3 per snippet                                             & \begin{tabular}[c]{@{}l@{}}200 snippets \\ of 5 turn dialog\end{tabular}                                                                 \\ \hline
\end{tabular}
\end{adjustbox}
\caption{Evaluators and samples in the surveyed papers}
\label{tab:evaluators-samples}
\end{table*}

\subsection{Sample sizes and annotators}

Table \ref{tab:evaluators-samples} shows the number of annotators and sample sizes used in the different papers. We have tried to do our best in collecting the information from the papers, however, these parameters were not always expressed clearly. The worst example is paper 3 that stated that they got multiple reviews, previews and feedback from the audience and the actors without specifying the exact number.

Most of the papers relied on non-expert annotators for conducting the evaluation with the exception of paper 1, 21 and 22, and partially paper 3. The use of experts is understandable as not just about anyone is competent enough to tell whether, for example, generated reviews for scientific papers (as in paper 22) are good or bad. However, this leads to a small number of evaluators as experts are difficult to recruit. Papers that did not use experts to evaluate the output either did not report any special requirements or mostly ensured that the evaluators were proficient enough in the language of the output.

In terms of the sample size, that is how many generated artefacts were evaluated, the amount varies a lot from anything starting from 2 as in paper 5 up to 250 as in papers 15 and 19. The samples were mostly picked at random, however some papers like paper 7 evaluated manually picked output.

There was also a lot of divergence in the number of annotators. Some papers had all annotators go through all samples like paper 21 and 22 did, while some other papers had several annotators that annotated the outputs so that each individual output was evaluated at least by 3 annotators like paper 14 and 23. Usually, there wasn't any clear discussion on how many outputs a given annotator annotated with the exception of paper 19, which reported that a given annotator could only annotate up to 30 outputs.

\subsection{Evaluation results}

An interesting point we wanted to pay attention to was the use of the evaluation results. After conducting a costly and time consuming human evaluation, one would hope that the results give a direction to the future research. However, this was not the case. All papers were limited to writing out the evaluation results and stating which system was better if the papers evaluated multiple systems. None of the papers was able to identify any concrete future directions for improving the generative system based on the human evaluation results. Human evaluation was merely there to provide some convincing evidence on the quality of the systems.

The only exception to this was paper 9. The authors conducted two different evaluations and they reached to an insightful conclusion. The two evaluation methods contradicted each other; according to the first evaluation, standard Finnish was preferred over dialectal one in all the parameters. However, the second evaluation showed that a dialectal poem was more often associated with originality, creativity and poem-likeness than its standard Finnish variant. The authors note that the results are not only dependent on how you conduct your human evaluation, but also on familiarity bias. In the first evaluation, where dialect was a controlled variable, the further the dialect was from standard Finnish, the lower it scored as the annotators were less familiar with it.

\section{Discussion}

There are currently many different creative NLG tasks people work with, and it is understandable that each task calls for slightly different evaluation methods. However, even work on closely related topics prefers to use their own evaluation methods that are not based on any existing research. And most alarmingly, if the evaluation is based on existing research, the evaluation questions are not motivated in the earlier research either. This type of evaluation has become to be known as a symptom of the Great Misalignment Problem \cite{hamalainen-alnajjar-2021-great}. When the evaluation is not targeted towards evaluating exactly what has been modelled, any type of evaluation that seems remotely related to the task becomes seemingly valid. 

However, when the evaluated parameters have only little to do with what was modelled, it is only evident that none of the surveyed papers was able to clearly identify the short-comings of their systems in such a way that they could propose some clear paths to follow for any future research. In fact, if you evaluate your system based on \textit{relatedness} and \textit{funniness} while neither is explicitly modelled, how can you know how to make your system more funny or produce more related output? The scores might have well been achieved by mere serendipity (the annotators happened to like the humor that happened to be in the small sample) (c.f. \citealt{gervas-2017-template}) or by data the model was trained on. 

Apart from the evaluation questions not aligning with the model, a much larger problem related to evaluation questions can be identified. Firstly, most of the papers were not clear about the actual evaluation questions used, instead they listed the evaluated parameters as though human evaluation was like an automated one where one can just score abstract notions such as \textit{typicality} or \textit{fluency} accurately on a 5 point scale. In other fields, it is known that even small changes in survey questions can lead to different survey results \cite{kalton1982effect,de2011framing,de2012effect}. Not revealing the actual questions only makes the situation worse. Another problem that rises from abstract evaluation questions is that it becomes less clear why the annotators gave certain answers.

Furthermore, people have a tendency on reading more into computer generated output than what the intention of the system was \cite{Veale+2016+73+92}. If you train a generative neural model on jokes, it will surely learn to output jokes, while it does not necessarily have any internal representation of humor. In such a case, the humor is purely in the eyes of the beholder and in the data the model was trained on, not in the method itself\footnote{See \citet{colton2008creativity} for discussion on the roles of the programmer, program and perceiver in creative systems}. For instance, \citet{alnajjar2019no} has shown that generated headlines were perceived more offensive by human annotators, while offensiveness was never modelled in the system.

While mostly every paper we surveyed opts for coming up with their own evaluation metrics, it is astonishing that these newly created evaluation settings are used as such. There are other fields dealing with human surveys that emphasize the need for conducting tests on your survey before conducting it in a larger scale to discover potential issues in your questionnaire \cite{collins2003pretesting,presser2004methods,thomas2004pilot}. None of the paper we surveyed discusses evaluation of evaluation. Instead, it is believed that any new evaluation metric the authors came up with just for a given paper will magically work as such and will yield scientifically valid results that will pass a peer review. All this while many of the papers ask questions using ambiguous terms such as \textit{fluency} (is something grammatical fluent? is something that seems to make sense semantically fluent? is something that is close to the annotator's own idiolect more fluent than something further away from it? is text generated in American English more fluent to Americans than text generated in British English? and so on) and \textit{coherence} (is something that repeats the same words coherent? can a complex figure of language be coherent if the annotator does not have time to think about it for more than a couple of seconds? does coherence have something to do with grammaticality as well? is a story that follows the same beliefs as the annotator seen as more coherent? and so on) that are reduced into a compact 1-5 scale that is later neatly averaged over all the annotators' opinions on all the samples. What does the average of 3.5 on a question all annotators might have interpreted differently even mean?

In other fields conducting online surveys, there are a lot of worries about selection bias of the human subjects \cite{bethlehem2010selection,greenacre2016importance}. This is hardly discussed in the fields of NLP and CC. Many of the papers we surveyed conducted their evaluation on a crowd-sourcing platform such as Amazon Mechanical Turk. None of the papers presented statistics on the demographics of the annotators. This might be a source of bias in the results. What makes such a bias even more problematic is the relatively small number of annotators that are usually recruited per individual output. Fields with more established human survey practices would not consider the typical 3-5 annotators of NLP and CC enough even for a \textit{qualitative} survey, which requires 5-25 participants \cite{creswell2016qualitative} or at least 6 participants \cite{morse1994designing}. However, human evaluation is usually conducted quantitatively, which means that the number of annotators depends heavily on multiple parameters and requires planning and justification on its own right \cite{bell1991big,lenth2001some,lavrakas2008encyclopedia}.

It is also very well known that people do not perceive things in a vacuum but rather as a continuum of stimuli where previously perceived stimuli affect to the next one. This effect is called priming (see \citealt{bloom2009encyclopedia}). To reduce the effect of priming or to have it consistent one should either shuffle the order in which the output is presented to the annotators or keep it always the same. Priming is especially in play in cases where annotators are to evaluate outputs produced by different systems. In such a case, output of a mediocre system might get greatly boosted when presented together with output by a bad systems. Nearly none of the papers we surveyed discussed this aspect of their evaluation setting.

Both CC and NLP have still a long way to go in order to reach to more sound human evaluation practices. However, INLG is still a step closer to scientific rigor as automated evaluation metrics were commonly used together with human evaluation, and sometimes as the only evaluation metric (such as \citealt{bien-etal-2020-recipenlg}), whereas ICCC had several papers presenting work on creative NLG without any evaluation at all (such as \citealt{agafonovaparanoid,petacpragmatics,wright2020creative}).

The use of experts in evaluation is something that should be taken under rigorous inspection in the future. Currently, there are contradicting studies on the topic indicating that consulting expert does have an effect in machine translation \cite{toral2018attaining} but not in poem generation \cite{lamb2017incorporating}. However, this is a question that is very likely to depend on the output that is to be evaluated and also on how the evaluation is conducted. 

Human computer interaction research has some more established methodologies for conducting human studies (see \citealt{jacko2012human,lazar2017research} such as cognitive walk-through (see \citealt{mahatody2010state}), human performance evaluation support system \cite{4395396} and user studies (see \citealt{mackenzie2015user}). These established methodologies could be taken into account when conducting evaluation of such an NLG system that calls for user interaction.

\section{Advices for future evaluation}

In this section, we outline how human evaluation of creative NLG systems should be conducted. We are not going to give an exact silver bullet framework to solve the problem, as the two fields are not at the state yet where enough would be known about human evaluation to state exactly how the evaluation needs to be conducted. Furthermore, we do not believe that a single fixed framework is enough to capture everything necessary in a topic as broad as creative text generation.

\subsection{Define the goals}

From the very early on, it is important to define what the goals of your system are (see \citealt{alnajjar2018master,jordanous2012standardised}). Try to be as concrete and precise as possible at this step. Once you have your goals clearly stated, it is easy to see the degree to which your implementation solution tries to achieve those goals and how much can be attributed to the method and how much to the training data. After this, the evaluation parameters will follow naturally from the goals you set for your system. This way, the evaluation questions do not appear seemingly from nowhere but are motivated by your research goals and implementation.

\subsection{Go concrete}

People have an inbuilt need to understand anything expressed in their language (see \citealt{Veale+2016+73+92}). This can lead easily into a situation, where annotators can read more into the evaluated output than what your system was aware of. By using evaluation questions that are as concrete as possible you can reduce the room for subjective interpretation (see \citealt{hamalainen2019let}). For example, for a pun like \textit{Becoming a vegetarian is a big missed steak} asking the annotators \textit{Is this humorous?} and \textit{Is this humorous because the pun "missed steak" sounds like "mistake"?} will result in different possible interpretations as the former question might let the annotators consider the generated joke funny for reasons other than those intended by the generative system.

\subsection{Run some tests}

As we have seen in this paper, the same concept can be evaluated through multiple different wordings and it is not always clear that the annotators understand the questions in the same way as the researchers intended. By running tests on your survey in real life, you can get more direct feedback than what you could get from annotators on Amazon Mechanical Turk. It is better to adjust your evaluation questions sooner than after running a costly crowd-sourcing.

Furthermore, the final number of annotators you need and how many samples you should evaluate depends on the evaluation task and setting. If you get high diversity in answers in the test run, you will probably need to have a larger number of annotators conducting the actual evaluation.

Testing is also a great way of seeing whether you are asking non-experts to evaluate things they consider too difficult or whether your questionnaire is too lengthy. You do not want your annotators to lose interest in the middle of the questionnaire and start annotating fast without paying too much attention.

\subsection{Run multiple evaluations}

Human evaluation does not need to be a one time thing conducted in a massive survey. You can run multiple different evaluations such as preference based ones, 5 point scale ones and true and false statements to better understand the limitations of your system and your human evaluation. The more evidence gathered by different evaluation methods you can show, the more confident you and other researchers can be of the quality of your method.

\subsection{Report everything clearly}

It is important to report the evaluation questions exactly as they were used, how the survey form was constructed including any instructions and wording used for the 5 point scale, and how the output was presented (always in the same order or shuffled). All these have an effect on the results. In software engineering, it is considered important to report any threats to the validity of the research \cite{feldt2010validity}. The same should apply to NLP and CC. One of the important threats to the validity of human evaluation is bias in the results. Therefore, it is important to report and discuss what kind of people participated in the evaluation survey.

\subsection{Analyze your results}

It is also important to dig deeper into the human evaluation results. If you as a researcher put a considerable amount of money in getting your human evaluation results, you should probably make the most out of them too. Instead of merely reporting the typical stats (mean, mode, median, standard deviation), why not looking into the best and worst performing output by the system as well and let the human evaluation be a guide in a deeper error analysis? This can open up insightful directions for future research. 

\section{Conclusions}

In this paper, we have surveyed papers presenting work on creative natural language generation that have been published in INLG 2020 and ICCC 2020. There have been many different evaluation methods including some unconventional ones such as critics' reviews and user testing. The most typical human evaluation method has been using a scaled survey, typically on a 5 point scale.

While most of the papers surveyed had come up with their own evaluation metrics, the most common parameters that have been evaluated were meaning, syntactic correctness, novelty, relevance and emotional value. Although, the terms used to refer to these notions have not been the same.

Most of the papers did not justify why they had evaluated certain parameters. Instead, the parameters were usually just stated as though they were an inarguable fact. It was more often than not the case that the actual evaluation questions were not revealed.

There was a lot of variation in the number of samples taken from the system output and how many annotators were used to conduct the evaluation. Typically the numbers were rather small. There was no discussion about the demographics of the annotators nor about what type of a bias it might have introduced.

Evaluation setups were never tested out beforehand, even though other fields dealing with human surveys recommend testing your questionnaires. This means that it is impossible to tell what the annotators really understood by the evaluation questions.

We established some advices for future evaluation, which include clearly defining the goal of the generative system, asking questions as concrete as possible, testing the evaluation setup, using multiple different evaluation setups, reporting the entire evaluation process and potential biases clearly, and finally analyzing the evaluation results in a more profound way than merely reporting the most typical statistics.

All in all, our fields, CC and NLP, have a lot to learn from other fields with longer traditions with human questionnaires in terms of conducting human evaluation. At the current stage, none of the papers we surveyed quite reached the same level of scientific rigor in their human evaluation as it is to be expected in other fields of science. However, this is not to say that the work of the authors of the papers we surveyed is inherently bad. This is just to highlight the fact that more attention needs to be paid in how human evaluation is conducted. Quite often with creative text generation, human judgment is the only viable metric to measure the performance of a system. Human evaluation of generated text has been conducted in the field of NLP already as early as in the 1960s \cite{mcdaniel-etal-1967-evaluation} it is a pity it has not caught up with the rest of the development in the field.

\bibliographystyle{acl_natbib}
\bibliography{acl2021}


\end{document}